\documentclass[runningheads]{llncs}
\usepackage{graphicx}
\usepackage{comment}
\usepackage{amsmath,amssymb} % define this before the line numbering.
\usepackage{color}

\usepackage{stmaryrd}

\usepackage[width=122mm,left=12mm,paperwidth=146mm,height=193mm,top=12mm,paperheight=217mm]{geometry}

\usepackage[table]{xcolor}
\def\best{\bf \cellcolor[gray]{0.85}}
\def\secbest{\cellcolor[gray]{0.92}}

\usepackage[caption=false]{subfig}
\usepackage{pgfplots}

\begin{document}
% \linenumbers
\pagestyle{headings}
\mainmatter
\def\ECCVSubNumber{1996}  % Insert your submission number here

\title{Improving Semantic Segmentation \\ via Self-Training} % Replace with your title

%\begin{comment}
% \titlerunning{ECCV-20 submission ID \ECCVSubNumber} 
% \authorrunning{ECCV-20 submission ID \ECCVSubNumber} 
% comment out the folllowing two lines
% \author{Anonymous ECCV submission}
% \institute{Paper ID \ECCVSubNumber}

\author{
Yi Zhu, Zhongyue Zhang, Chongruo Wu\thanks{\scriptsize{Work done during an internship at Amazon.}}, Zhi Zhang, Tong He,\\ Hang Zhang, R. Manmatha, Mu Li, Alexander Smola
}

\institute{Amazon Web Services \hspace{2mm} 
University of California, Davis\footnotemark[1]\\
\email{\{yzaws,zhongyue,chongrwu,zhiz,htong,hzaws,manmatha,mli,smola\}@amazon.com}}

% % %\affil[ ]{Amazon Web Services}
% % \affil[ ]{ 
% % \tt\small \{yzaws,zhongyue,chongrwu,zhiz,hzaws}
% % \affil[ ]{\tt\small  htong,jonasmue,manmatha,mli,smola\}@amazon.com}

% %\end{comment}
%******************


\titlerunning{Improving Semantic Segmentation via Self-Training}
% \author{
% Yi Zhu, Zhongyue Zhang, Chongruo Wu, Zhi Zhang, Hang Zhang, \\ Tong He, R. Manmatha, Mu Li
% }

% \institute{Amazon Web Services \\
% \email{\{yzaws,zhongyue,chongrwu,zhiz,hzaws,htong,manmatha,mli\}@amazon.com}}

%
\authorrunning{Yi Zhu et al.}
\begin{comment}
\titlerunning{ECCV-20 submission ID \ECCVSubNumber} 
\authorrunning{ECCV-20 submission ID \ECCVSubNumber} 
\author{Anonymous ECCV submission}
\institute{Paper ID \ECCVSubNumber}
\end{comment}
%******************
\maketitle

\begin{abstract}
Deep learning usually achieves the best results with complete supervision. In the case of semantic segmentation, this means that large amounts of pixelwise annotations are required to learn accurate models. In this paper, we show that we can obtain state-of-the-art results using a semi-supervised approach, specifically a self-training paradigm. We first train a teacher model on labeled data, and then generate pseudo labels on a large set of unlabeled data. Our robust training framework can digest human-annotated and pseudo labels jointly and achieve top performances on Cityscapes, CamVid and KITTI datasets while requiring significantly less supervision. We also demonstrate the effectiveness of self-training on a challenging cross-domain generalization task, outperforming conventional finetuning method by a large margin.
Lastly, to alleviate the computational burden caused by the large amount of pseudo labels, we propose a fast training schedule to accelerate the training of segmentation models by up to 2x without performance degradation.

\end{abstract}

\section{Introduction}
\label{sec:introduction}
Semantic segmentation is a fundamental computer vision task whose goal is to predict semantic labels for each pixel. Great progress has been made in the last few years in part thanks to the collection of large and rich datasets with high quality human-annotated labels. However, pixel-by-pixel annotation is prohibitively expensive, e.g., labeling all pixels in one Cityscapes image takes more than an hour \cite{Cordts2016Cityscapes}. Thus until now, we only have semantic segmentation datasets consisting of thousands or tens of thousands annotated images \cite{Cordts2016Cityscapes,Neuhold2017mapillaryVista,BDD100K}, which are orders of magnitude smaller than datasets in other domains \cite{Sun2017JFT300M,Mahajan2018WeakSupLimit}. 

Given the fact that learning with annotated samples alone is neither scalable nor generalizable, there is a surge of interest in using unlabeled data by semi-supervised learning. \cite{Souly2017SemiGAN,Hung2018Adversarial} adopt the concept of adversarial learning to improve a segmentation model.
Self-training \cite{Luc2017futureSeg,Zou2018DAClassBalance,Li_2019_bidirection,Zou_2019_CRST,Lian_2019_Pyramid} often uses a teacher model to generate extra annotations from unlabeled images. 
Recently, \cite{Mustikovela2016labelPropagation,Budvytis2017augmentation} use temporal consistency constraints to propagate ground truth labels to unlabeled video frames. However, these models often beat a competing baseline in their own settings but have difficulty achieving state-of-the-art semantic segmentation performance on widely adopted benchmark datasets.

\begin{figure}[t]
\centering
\includegraphics[width=1.0\linewidth]{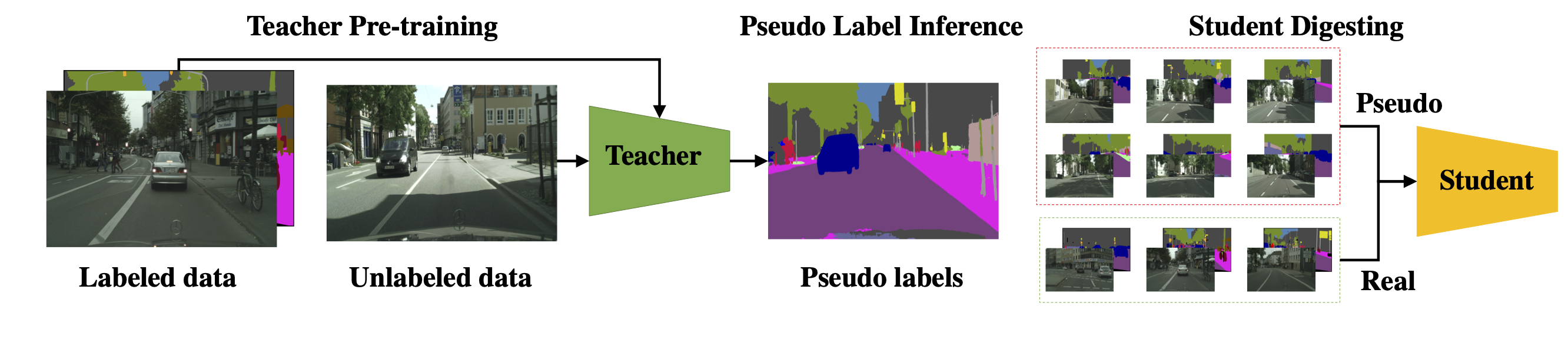}
\vspace{-4ex}
\caption{Overview of our self-training framework. (a) Train a teacher model on labeled data. (b) Generate pseudo labels on a large set of unlabeled data. (c) Train a student model using the combination of both real and pseudo labels}
\label{fig:self_training}
\vspace{-4ex}
\end{figure}

In this paper, we would like to revisit semi-supervised learning in semantic segmentation, particularly on driving scene segmentation. Driving scene segmentation suffers from insufficient training labels but it has access to unlimited unlabeled images collected by running vehicles. Semi-supervised learning is thus perfectly suited for such a task. 
% for a wide range of driving scenes
In addition, motivated by an open problem in semantic segmentation, i.e., a segmentation model trained on one driving dataset may not generalize to another due to domain gap, we also evaluate our model on a challenging cross-domain generalization task (e.g., from Cityscapes to Mapillary) with a small number of annotations in the target domain.

The method we used is based on the self-training framework \cite{Xie2019NoisyStudent}, which is illustrated in Fig.~\ref{fig:self_training}. We first train a \textit{teacher model} on a small set of labeled data, which we refer as ``real labels''. We then generate ``pseudo labels'' by using this teacher model to predict on a large set of unlabeled data. In the end, a \textit{student model} is trained using the combination of both real labels and pseudo labels. If the noise of the pseudo labels can be properly handled, the student model often outperforms the teacher model. And in particular, when the pseudo labels come from a different domain, the student model should generalize better than the teacher model on this new domain. 

As we can generate unlimited pseudo labels, it significantly increases the training computational cost. A key reason segmentation models are hard to train is because of the high resolution input images. We propose a schedule that reduces the image resolution during training from time to time. A small resolution reduces both the computation cost per image and also allows a large batch for a better system efficiency. We carefully design the schedule so that it will not impact the final performance given a fixed number of images processed.

Extensive experimental results demonstrate the effectiveness of our approach on three driving scene segmentation datasets \cite{Cordts2016Cityscapes,Brostow2008camvid,Geiger2012CVPR}. Take Cityscapes \cite{Cordts2016Cityscapes} as an example, we achieve an mIoU of $82.7\%$ on its test set using only fine annotations. We outperform all prior arts that use the same training data, and several methods \cite{Chen2018SearchSegDPC,Liu2019autodeeplab,Bulo2018inplaceABN,Zhu2019VPLR} trained using additional labeled data.

Our contributions are summarized below:
\begin{itemize}
    \item We introduce a self-training framework for semantic segmentation, which is able to properly handle the noisy pixel-level pseudo labels. 
    
    \item We propose a training schedule that adjusts the image resolution during the training to speedup the performance without losing model accuracy. 
    
    \item We extensively evaluate the proposed method to show that it 
    achieves state-of-the-art performance on Cityscapes, CamVid and KITTI datasets, while using significantly less annotations.
    
    \item We demonstrate the effectiveness of self-training on a challenging cross-domain generalization task with different semantic categories.
    
\end{itemize}

\section{Related Work}
\label{sec:related_work}

\textbf{Semantic segmentation.} Starting from the seminal work of FCN \cite{Long2015FCN}, semantic segmentation has made significant progress \cite{Zhao2017pspnet,Chen2018deeplabv3plus,Bulo2018inplaceABN,zhangli_dgcn,Cheng2019DeepLabPano} with respect to model development. Recent work has focused more on exploiting object context by using attention \cite{Yuan2019OCNetv2,Fu2018DANet,Zhang_encnet_CVPR18,Fu2019ACNet,zhao2018psanet,Zhang2019ACFNet}, designing more efficient networks \cite{Huang2019CCNet,Poudel2019FastSCNN,Yu_BiSeNet_2018eccv,Li2019DFANet,wu2019fastfcn} and performing neural architecture search \cite{Liu2019autodeeplab,Chen2018SearchSegDPC,Nekrasov_2019_fastNASSeg,Zhang_2019_CNAS}, etc.

Our work is different as we introduce a self-training framework for semantic segmentation to explore the benefit of using unlabeled data. We demonstrate that our method is orthogonal to model development. We can improve a number of widely adopted models irrespective of the network architectures. 

\noindent \textbf{Semi-supervised learning.} Recently we have witnessed the power of semi-supervised learning in the image classification domain. By using a large amount of unlabeled data, \cite{Yalniz2019BillionSemi,Xie2019NoisyStudent} are able to achieve state-of-the-art performance on ImageNet. There are also numerous papers on semi-supervised semantic segmentation, such as using adversarial learning \cite{Souly2017SemiGAN,Hung2018Adversarial}, self-training \cite{Luc2017futureSeg,Zou2018DAClassBalance,Li_2019_bidirection,Zou_2019_CRST,Lian_2019_Pyramid}, consistency regularization \cite{Mittal2019HighLow,French2019Perturbations}, knowledge distillation \cite{Xie2018TeacherStudent,Liu2019KDSegmentation}, video label propagation \cite{Mustikovela2016labelPropagation,Budvytis2017augmentation}, etc. However, these models often beat a competing baseline in their own settings but have difficulty achieving state-of-the-art semantic segmentation performance on widely adopted benchmark datasets.
 
In this paper, we revisit semi-supervised learning for semantic segmentation. Our work is different from the previous self-training literature \cite{Luc2017futureSeg,Zou2018DAClassBalance,Li_2019_bidirection,Zou_2019_CRST,Lian_2019_Pyramid} because our problem settings are different, and no direct comparisons can be made. Besides, we adopt the teacher-student paradigm without using additional regularization as in \cite{Zou2018DAClassBalance,Zou_2019_CRST,Lian_2019_Pyramid}.
Our work also differs from knowledge distillation \cite{Xie2018TeacherStudent,Liu2019KDSegmentation} since our goal is to get state-of-the-art performance by using unlabeled data, not distilling a light-weight student model for fast semantic segmentation. 

\noindent \textbf{Domain adaptation.} Our cross-domain generalization task is related to unsupervised domain adaptation (UDA).
% in terms of having access to both images and labels in the source domain and images alone in the target domain.
The goal of UDA is to learn a domain invariant feature representation to address the domain gap/shift problem.
Numerous approaches have been proposed in this field, such as the dominating adversarial learning pipeline \cite{Hoffman2018CyCADA,Tsai_adaptseg_2018,Vu_2019_CVPR,Luo_2019_SAIB,zhang2019category,Yang2020LabelRecons}, conservative loss \cite{Zhu_2018_ECCV}, texture/structure invariant \cite{Chen_2019_geometric,chang2020texture}, consistency regularization \cite{Chen_2019_CrDoCo,Lee_2019_SWD,Yue_2019_DRPC}, etc. 

Our work differs from conventional UDA in multiple ways. First, our settings are different. Most UDA literature adopt a synthetic-to-real setting (e.g., GTA5/SYNTHIA to Cityscapes), while we consider a real-to-real setting (e.g., Cityscapes to Mapillary/BDD100K). Second, our framework is more generic and simpler. We do not use adversarial learning or specially designed loss functions to reduce the domain gap, and yet achieve decent generalization performance. Third, we show promising results on a challenging but practical task where the target domain has new classes and we have a few labeled samples. Most UDA literature do not consider this scenario. They often assume there is no labeled data in the target domain and only report performance when the source and target domain have the same number of classes.

\vspace{-2ex}
\section{Methodology}
\label{sec:methodology}

In this section, we introduce our self-training method for semantic segmentation. We first describe the employed teacher-student framework in Sec. \ref{subsec:self_training}. With the large amount of pseudo labels, we propose a centroid data sampling technique in Sec. \ref{subsec:centroid_sampling} to combat the class imbalance problem and the noisy label problem. Following this, we design a fast training schedule in Sec. \ref{subsec:fast_training} to handle the large expanded training set. This approach speeds up the model training by up to 2x without performance degradation. Finally, we demonstrate in Sec. \ref{subsec:cross_domain} that our self-training method can also greatly improve the performance on a challenging cross-domain generalization task with only a few labels.

\vspace{-2ex}
\subsection{Self-Training using Unlabeled Data}
\label{subsec:self_training}
While semi-supervised learning has been widely studied for semantic segmentation \cite{Luc2017futureSeg,Zou2018DAClassBalance,Mittal2019HighLow,French2019Perturbations,Xie2018TeacherStudent,Liu2019KDSegmentation}, it is difficult to beat human supervised counterparts \cite{Cheng2019DeepLabPano,Zhu2019VPLR,Li2019GALDNet,Yuan2019OCNetv2}. In this section, we introduce a teacher-student framework to perform self-training on semantic segmentation. Our goal is to use a small set of labeled data and a large quantity of unlabeled data to improve both the accuracy and robustness of semantic segmentation models. In this way, the human labeling effort can be largely reduced. 

We present an overview of our self-training framework in Fig. \ref{fig:self_training}. Given a small quantity of labeled training samples (an image and a human-annotated segmentation mask), we first train a teacher model with standard cross-entropy loss. We adopt a number of training techniques specially designed for semantic segmentation, to make the teacher model as good as possible. We will discuss the training details in the following sections. 

We then use the teacher model to generate pseudo labels on a large number of unlabeled images. The better the teacher model is, the higher the quality of the generated pseudo labels can be. As can be seen in Fig. \ref{fig:self_training}, our teacher-generated pseudo labels have a good quality that are close to human annotations. More visualizations of pseudo labels can be found in the Appendix. 

Finally, we train a student model using both human-annotated labels (real labels) and teacher-generated labels (pseudo labels). 
Note that the teacher-student framework is widely studied in the literature of distillation, however, it has been reported in~\cite{He2019BagTricks} to have the limitation that teacher and student are expected to have a similar architecture to work well. 
We would like to point out that with our approach, the student model may have a different network architecture which does not have to be the same as that of the teacher's. In our experiments, we use a single teacher but train several student models with various backbones and network architectures. Our self-training framework consistently improves all of them, which demonstrate its generalizability.

\vspace{-2ex}
\subsection{Fighting Class Imbalance with Centroid Sampling}
\label{subsec:centroid_sampling}
The biggest challenge in self-training is how to deal with the noise in pseudo labels \cite{Yalniz2019BillionSemi,Xie2019NoisyStudent,Zou2018DAClassBalance,Li_2019_bidirection,Zou_2019_CRST,Lian_2019_Pyramid}. With the increased number of training samples, there is a high chance that noisy samples mislead and confuse the model during training. In addition, semantic segmentation is a dense prediction problem and  each erroneous pixel prediction is a noisy sample. That explains why previous self-training papers \cite{Mustikovela2016labelPropagation,Luc2017futureSeg} do not observe significant improvement even though they generate a massive amount of pseudo labels from unlabeled images.

There are several widely adopted approaches to control the usage of pseudo samples, such as (1) lowering the ratio of pseudo labels in each mini-batch (or in each training epoch); (2) selecting pseudo labels with high confidence; (3) setting lower weights in computing the loss for pseudo labels. However, these methods do not account for the problem of class imbalance. We argue that this problem will become more severe in the self-training paradigm because samples of each class are amplified by a biased teacher model with the increasing number of unlabeled training samples. For example, in the original Cityscapes training set, the ``road'' class has 360 times more pixels than the ``motorcycle'' class. In the expanded training set with pseudo labels, this ratio is on the scale of thousands. Hence, a better way to control the usage of pseudo labels is essential to make self-training work in the semantic segmentation domain. 

We introduce a centroid sampling strategy similar to \cite{Bulo2018inplaceABN,Zhu2019VPLR} but with several major differences. The idea is to make sure that our model can see instances from all classes in the expanded noisy set within each epoch, even for the underrepresented classes. To be specific, we first record the centroid of areas containing the class of interest before training. A centroid is the arithmetic mean position of all the points within an object. For example, the centroid of a car instance will be a point roughly in the middle of the car. Then during training, we can query training samples using such class-level information, i.e., crop an image patch around the centroid. This centroid sampling strategy has two advantages. First, we can train a semantic segmentation model without worrying about the problem of class imbalance, no matter how large the noisy set is. We can choose to uniformly query training mini-batches using the centroids so that the model can see instances from all classes. 
Second, it does not break the underlying data distribution because we only crop the image around the centroids. The ratio of pixel count among all classes will remain approximately the same.

We now illustrate the differences between our centroid sampling technique and \cite{Zhu2019VPLR}. First, \cite{Zhu2019VPLR} fixes the number of iterations within each epoch, which means the model can only see a small fraction of all the pseudo labels. This is not suitable for the self-training paradigm, because we can easily generate a huge number of pseudo labels. We relax this constraint to flexibly adjust the ratio between real and pseudo labels in each epoch. In this way, we can fully understand the effect brought by the pseudo labels, and determine the appropriate amount of pseudo labels to train a good student model. 
Second, \cite{Zhu2019VPLR} only selects underrepresented classes such as ``fence'', ``rider'', ``train'' to augment the original dataset. We instead use all the classes because we believe it is better to train the model following the real-world data distribution. It is not necessary to force class ``road'' and class ``motorcycle'' to have the same amount of pixels.

\vspace{-2ex}
\subsection{Fast Training Schedule for Large Expanded Dataset}
\label{subsec:fast_training}
Once we have the pseudo labels and a principled way to control the noisiness, it is time to kickoff training. However, semantic segmentation consumes lots of computing resources. Due to the large crop size (e.g., $800 \times 800$), we can only use a small batch size (2 or 4) depending on the network architecture to fit into the GPU memory. Hence, even training on a medium-scale dataset like Cityscapes with only 3K training samples using a high-end 8-GPU machine, takes days to finish. Now if we increase the size of dataset to 10 times more, the training will take weeks to complete, which leads to long research cycles. This is part of the reason why self-training hasn't been investigated thoroughly in the field of semantic segmentation. 

Given that the slowness is caused by using a small batch size, can we reduce the crop size during training to increase the batch size? Several researchers \cite{Li2019GALDNet,Zhao2017pspnet,Chen2018deeplabv3plus} have done this experiment to accelerate their model training, and they came to the conclusion that reducing crop size hurts the results. Thus, trading crop size for batch size is not worth it. We agree with the conclusion because semantic segmentation is a per-pixel dense prediction problem and a small crop size will lead to the problem of losing global context and detailed boundary information, which is of course harmful. However, what if we can design a training schedule that iterates between a small crop size and a larger crop size, so that the training time can be reduced without losing segmentation accuracy?

\begin{figure}[t]
\centering
\includegraphics[scale=0.423]{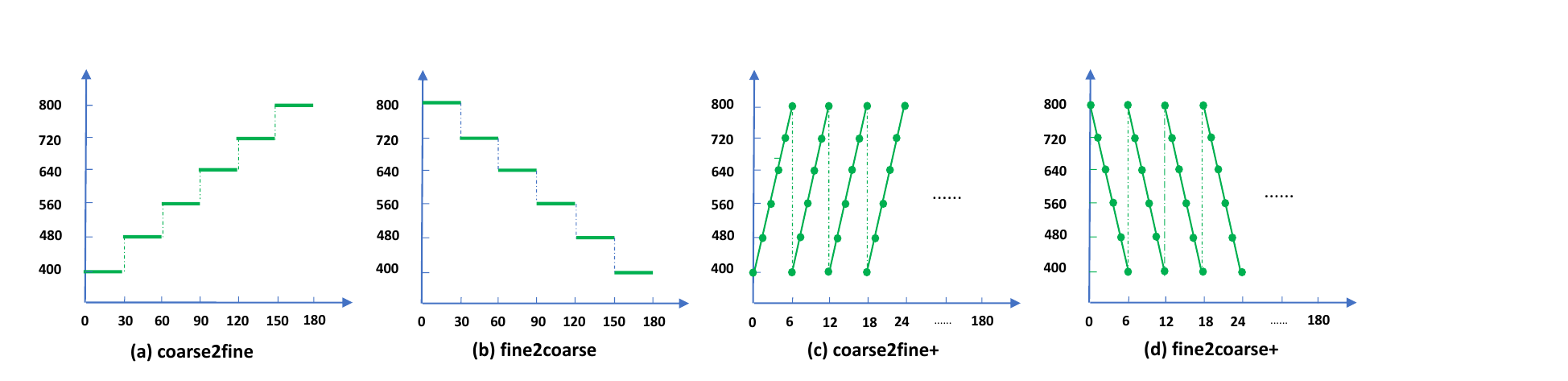}
\vspace{-4ex}
\caption{Overview of our proposed fast training schedules. x-axis is the epoch number and y-axis is the crop size. See texts in Section \ref{subsec:fast_training} for more details}
\label{fig:fast_training}
\vspace{-4ex}
\end{figure}

In this work, inspired by the coarse-to-fine method in computer vision \cite{Wu2020multigrid}, we design several training schedules to speed up the experiments.
Our goal is to avoid the speed and accuracy trade-off, and achieve faster training without losing accuracy. As shown in Fig. \ref{fig:fast_training}, we introduce 4 learning schedules, namely coarse2fine, fine2coarse, coarse2fine+ and fine2coarse+. 
To be specific, (1) coarse2fine means we first use small crop sizes such as 400 in the early epochs, then change to 480, 560, 640, 720 and eventually 800 for the rest of the training. Each crop size stays constant for several epochs, e.g., we change the crop size every 30 epochs. (2) fine2coarse means we first use a large crop size such as 800 in the early epochs, then change to 720, 640, 560, 480 and 400 as we progress to the end. (3) coarse2fine+ means we iterate the crop size every epoch to maximize the scale variation during model learning. For example, we use crop size 400 for epoch 0, 480 for epoch 1, 560 for epoch 2, and so on. And similarly, (4) fine2coarse+ will be the reverse process of (3) where we start with a larger (finer) crop. We will show the performance of each learning schedule in Section \ref{sec:experiments}, and demonstrate that the coarse2fine+ schedule is the best candidate. It is able to speed up the training by up to 2x with the same segmentation accuracy. 

We would like to point out that our fast training schedule is a general technique. It is not only suitable for our self-training framework with large expanded training set, but also applicable to standard semantic segmentation training on large-scale datasets such as Mapillary \cite{Neuhold2017mapillaryVista} and BDD100K \cite{BDD100K}.
% or ApolloScape \cite{apolloscape_arXiv_2018}.

\vspace{-2ex}
\subsection{Cross-Domain Generalization with New Categories}
\label{subsec:cross_domain}

Using the self-training framework, we can improve the accuracy of semantic segmentation on the current dataset by leveraging a large set of external unlabeled data. However, a segmentation model trained on one dataset may not generalize to another. For instance, two driving datasets collected in different locations may be significantly different in terms of traffic, lighting and viewpoint. On the other hand, in practice it is a meaningful task as people usually train model on one mature dataset with abundant labels, and expect to test on another new dataset with a small number of annotations. Note that the target dataset could have new categories which makes the task even more difficult, as the model needs to learn knowledge from new scenarios with just a few labels.

A conventional approach would be training a model on the source dataset, and finetuning it on the new dataset assuming the two datasets share similar distribution. However, this approach requires a large amount of annotations from the target dataset to achieve good performance. Here we introduce the improvement with our self-training framework for such a challenging setting. 

We first train a good model on the source domain, using both real labels from the source domain and pseudo labels from the target domain. In this manner, our model learns the prior knowledge of the data distribution in the target domain, even though information about the new semantic categories is not provided. Then we finetune this model on the small set of labeled samples from the target domain, so that it can quickly adapt to the new domain. In our experiments, we have shown that our self-training framework can handle the problem of cross-domain generalization with new categories better than conventional finetuning approach. Even when there are just 10 annotations per category, our method is able to achieve decent segmentation accuracy.

\section{Experiments}
\label{sec:experiments}
\vspace{-1ex}
In this section, we describe the implementation details of our framework. Then we report the performance on three widely adopted driving scene segmentation datasets, Cityscapes \cite{Cordts2016Cityscapes}, CamVid \cite{Brostow2008camvid} and KITTI \cite{Geiger2012CVPR}. We perform all the ablation studies on Cityscapes because it is the most benchmarked dataset. In the end, we evaluate our model on a cross-domain generalization task from Cityscapes to Mapillary. For all the datasets, we use the standard mean Intersection over Union (mIoU) metric to report segmentation accuracy.  

\setlength{\tabcolsep}{4pt}
\begin{table}[t]
\begin{center}
\caption{Performance of our baseline and the results of self-training. Fine: using Cityscapes fine annotations. Coarse$^{*}$ and Mapillary$^{*}$ means we only use the images from Cityscapes coarse dataset and Mapillary dataset, without using their labels}
\label{table:baseline}
\resizebox{0.4\columnwidth}{!}{%
\begin{tabular}{l  c c c c}
\hline\noalign{\smallskip}
 & Fine & Coarse$^{*}$ & Mapillary$^{*}$ & mIoU ($\%$)\\
\noalign{\smallskip}
\hline
\noalign{\smallskip}
% Vanilla  & $\checkmark$ & & & & & 76.6 \\
%   & $\checkmark$ & $\checkmark$& & & & 77.5 \\
Teacher  & $\checkmark$ & &  & 78.1 \\
\noalign{\smallskip}
\hline
\noalign{\smallskip}
Student  & $\checkmark$ & $\checkmark$  &  & 79.0 \\
Student  & $\checkmark$ & $\checkmark$ & $\checkmark$ & \textbf{79.3} \\
% \noalign{\smallskip}
% \hline
% \noalign{\smallskip}
% Oracle  & $\checkmark$ & & &$\checkmark$ & $\checkmark$ & 79.8 \\
\hline
\end{tabular}
}
\end{center}
\vspace{-4ex}
\end{table}
\setlength{\tabcolsep}{1.4pt}

\vspace{-1ex}
\subsection{Datasets}
\label{subsec:datasets}

\textbf{Cityscapes} contains 5K high quality annotated images, splitting into $2975$ training, $500$ validation, and $1525$ test images. The dataset defines $19$ semantic labels and a background class. There are also $20$K coarsely annotated images, but we ignore their labels in our self-training framework.  \textbf{KITTI} has the same data format and metrics with Cityscapes, but with varying image resolution. The dataset consists of 200 training and 200 test images, without an official validation set. \textbf{CamVid} defines $32$ semantic labels, however, most literature only focuses on 11 of them. It includes 701 densely annotated images, splitting into 367 training, 101 validation and 233 test images. \textbf{Mapillary Vista} is a recent large-scale benchmark with global reach and includes more varieties. The dataset has 18K training, 2K validation and  5K test images. We use its research edition which contains 66 classes. We will refer to it as Mapillary in the paper.

\vspace{-1ex}
\subsection{Implementation Details}
\label{subsec:implementation}
We employ the SGD optimizer for all the experiments. We set the initial learning rate to $0.02$ for training from scratch and $0.002$ for finetuning. We use a polynomial learning rate policy \cite{Liu2017parsenet}, where the initial learning rate is multiplied by $(1 - \frac{\text{epoch}}{\text{max}\_\text{epoch}})^\text{power}$ with a power of $0.9$. Momentum and weight decay are set to $0.9$ and $0.0001$ respectively. Synchronized batch normalization \cite{Zhao2017pspnet} is used with a default batch size of 16. Using our fast training schedule, the batch size can increase to $64$ when the crop size is smaller. The number of training epochs is set to $180$ for both Cityscapes and Mapillary, $80$ for CamVid and $50$ for KITTI. The crop size is set to $800$ for both Cityscapes and Mapillary, $640$ for CamVid and $368$ for KITTI due to different image resolutions. For data augmentation, we perform random spatial scaling (from 0.5 to 2.0), horizontal flipping, Gaussian blur and color jittering (0.1) during training. We adopt DeepLabV3+ \cite{Chen2018deeplabv3plus} as our network architecture, and use ResNeXt50 \cite{Xie2017ResNeXt} as the backbone for the ablation studies, and WideResNet38 \cite{Wu2016WideOrDeep} for the final test-submissions. We adopt the OHEM loss following \cite{Wu2016OHEMSeg,Yuan2019OCNetv2}.
% where the hard pixels are defined as having probabilities smaller than $\theta$ over the correct classes and a mini-batch of at least $K$ pixels. We set $\theta$ to $0.7$ and $K$ to $100000$. 
For all the ablation experiments, we run the same training recipe five times and report the average mIoU. 
% For our centroid sampling, we record the centroid of each class instance with a shape of 1024 $\times$ 1024 because 1024 is the shorter side of Cityscapes images.

\setlength{\tabcolsep}{4pt}
\begin{table}[t]
\begin{center}
\caption{Ablation study on the ratio between pseudo and real labels. We find that increasing the ratio of pseudo labels improves the segmentation accuracy, and even outperforms a model pre-trained on 43K real labels \cite{Zhu2019VPLR}. CS: centroid sampling}
\label{table:ratio}
\begin{minipage}{0.6\textwidth}%
	\centering
	\subfloat{(a) Self-training with pseudo labels}{
		\scalebox{0.8}{
			\begin{tabular}{l c c c c}
                \hline\noalign{\smallskip}
                & Real & Pseudo & w/o CS & w CS\\
                \noalign{\smallskip}
                \hline
                \noalign{\smallskip}
                Teacher   & 3K & - & 78.1 & 78.1 \\
                \noalign{\smallskip}
                \hline
                \noalign{\smallskip}
                Student   & 1.5K & 1.5K & 78.4 & 79.3  \\
                Student   & 1.5K & 4.5K & 78.7 & 79.7 \\
                Student   & 1.5K & 7.5K & 78.9 & 79.9  \\
                Student   & 1.5K & 10.5K & 78.7 & \textbf{80.0} \\
                \noalign{\smallskip}
                \hline
                \noalign{\smallskip}
                VPLR \cite{Zhu2019VPLR}  & 43K  & - & - & 79.8 \\
                \hline
            \end{tabular}
		}
	}
\end{minipage}%
%\qquad
\begin{minipage}{0.4\textwidth}%
	\centering
	\subfloat{(b) Duplicate real labels}{
% 	\vspace{9ex}
		\scalebox{0.8}{
			\begin{tabular}{l  c c c}
                \hline\noalign{\smallskip}
                &  Real & mIoU ($\%$)\\
                \noalign{\smallskip}
                \hline
                \noalign{\smallskip}
                Teacher   & 3K & 78.1 \\
                Teacher   & 6K & 78.9 \\
                Teacher   & 9K & \textbf{79.1} \\
                Teacher   & 12K & 79.0 \\
                \hline
            \end{tabular}
		}
	}
\end{minipage}
\end{center}
\vspace{-6ex}
\end{table} 
\setlength{\tabcolsep}{4pt}

\subsection{Cityscapes}
\label{subsec:cityscapes}

We first perform ablation studies on the validation set of Cityscapes to justify our framework design and then report our performance on its test set.

\noindent \textbf{Self-training baseline}
Here we establish the baseline for all of our experiments that follow. As shown in the top of Table \ref{table:baseline}, our baseline model (DeepLabV3+ with ResNeXt50 backbone using OHEM loss) trained on the 3K Cityscapes fine annotations has an mIoU of $78.1\%$ \cite{Zhu2019VPLR}. We adopt this model as the teacher to start self-training. 

Next, we use this model to generate pseudo labels on the Cityscapes coarse images and Mapillary images. Note that the images in both the Cityscapes coarse dataset and Mapillary dataset are annotated but we ignore the labels and treat them as unlabeled data. Since both new datasets have 20K pseudo labels, we randomly pick half the training samples (1.5K) from the Cityscapes fine annotations, and another half (1.5K) from the centroids of generated pseudo labels for a fair comparison. It has been shown that longer training is beneficial for dense prediction tasks such as semantic segmentation because it can refine the boundaries \cite{Poudel2019FastSCNN,Cheng2019DeepLabPano,Li2019GALDNet}. In this way, the total number of training samples (3K) within an epoch remains the same with our baseline.

As shown in the bottom of Table \ref{table:baseline}, with the pseudo labels generated on the Cityscapes coarse images, it brings $0.9\%$ mIoU improvement ($78.1\% \shortrightarrow 79.0\%$). Together with the pseudo labels generated on the Mapillary images, we obtain an $1.2\%$ mIoU improvement over the baseline ($78.1\% \shortrightarrow 79.3\%$). 
Our preliminary results suggest the potential of self-training for semantic segmentation. We can improve a strong baseline without using extra labeled data. Since adding pseudo labels from both Citysapes coarse and Mapillary gives the best result, we will use them for the rest of our experiments, which means the amount of our pseudo labels is 40K. Again for clarity, we will term the 3K Cityscapes fine annotations as our real labels, and the 40K teacher-generated labels as pseudo labels afterwards.

\setlength{\tabcolsep}{5pt}
\begin{table}[t]
\begin{center}
\caption{Our self-training method can improve the student models irrespective of backbones and network architectures. We again outperform models pre-trained on Mapillary labeled data for all three students. $\#$ Real: the number of real labels used in training}
\label{table:otherstudents}
\resizebox{0.8\columnwidth}{!}{%
\begin{tabular}{l l l c c}
\hline\noalign{\smallskip}
 & Network & Backbone & $\#$ Real & mIoU ($\%$)\\
\noalign{\smallskip}
\hline
\noalign{\smallskip}
Teacher & DeepLabv3+  & ResNeXt50  & 3K & 78.1 \\
% \noalign{\smallskip}
% \hline
% \noalign{\smallskip}
% Baseline   & DeepLabV3+  & ResNeXt50  & 78.1 \\
% Pre-trained \cite{Zhu2019VPLR}   & DeepLabV3+  & ResNeXt50  & 79.8 \\
% Self-training   & DeepLabV3+  & ResNeXt50  & \textbf{80.0} \\
\noalign{\smallskip}
\hline
\noalign{\smallskip}
Baseline \cite{Zhu2019VPLR}   & DeepLabV3+  & WideResNet38  & 3K & 80.5\\
Mapillary Pre-trained  \cite{Zhu2019VPLR}    & DeepLabV3+  & WideResNet38  & 43K & 81.5 \\
Self-training(ours)   & DeepLabV3+  & WideResNet38  & 3K & \textbf{82.2} \\
\noalign{\smallskip}
\hline
\noalign{\smallskip}
Baseline \cite{Poudel2019FastSCNN}  & FastSCNN  &  - & 3K & 68.6 \\
Mapillary Pre-trained \cite{Poudel2019FastSCNN}    & FastSCNN  & -  & 43K &  71.7 \\
Self-training(ours)   & FastSCNN  &  - & 3K & \textbf{72.5} \\
\noalign{\smallskip}
\hline
\noalign{\smallskip}
Baseline \cite{Zhao2017pspnet}  & PSPNet  &  ResNet101 & 3K & 77.9 \\
Mapillary Pre-trained  \cite{Zhao2017pspnet}    & PSPNet  & ResNet101  &  43K & 79.2 \\
Self-training(ours)   & PSPNet  &  ResNet101 & 3K & \textbf{79.9} \\
\hline
\end{tabular}
}
\end{center}
\vspace{-6ex}
\end{table}

\noindent \textbf{Ratio of pseudo labels to real labels}
In the previous experiment, we only pick 1.5K samples randomly from the pool of 40K pseudo labels, thus we may miss most of our teacher-generated data and cannot fully explore the potential of self-training.
Hence, we would like to increase the ratio between pseudo labels and real labels from 1:1 to 3:1, 5:1 and 7:1. Specifically, for each ratio setting we still randomly pick 1.5K real labels and pick 1.5K, 4.5K, 7.5K and 10.5K samples from the pool of pseudo labels. Under each setting, we also compare training with and without the centroid sampling. The experimental results are reported in Table \ref{table:ratio}a. To cancel out the effect from longer training, we design another group of baseline experiments by duplicating the real labels. For example, a ratio of 3:1 between pseudo and real labels equals 6K training samples, whose baseline counterpart is obtained by duplicating the Cityscapes training set twice. The experimental results are reported in Table \ref{table:ratio}b.

As we can see in Table \ref{table:ratio}a, increasing the ratio indeed improves the segmentation accuracy, from $79.3\%$ to $80.0\%$. We also show that centroid sampling is essential to achieving good results. Without it, our model trained on the same amount of pseudo labels can only reach an mIoU of $78.9\%$, which is $1.1\%$ worse than using centroid sampling. Furthermore, in the situation that pseudo labels dominate the training set (e.g., 10.5K samples), centroid sampling does a good job in controlling label noisiness, otherwise the performance starts to drop ($78.9\% \shortrightarrow 78.7\%$).
% This indicates that more pseudo labels will help model training, serving as another form of augmentation to achieve better and more robustness. 
One observation we want to point out is, our final performance of $80.0$ is even better than a model pre-trained on Mapillary labeled data and finetuned on both Cityscapes fine and coarse annotations \cite{Zhu2019VPLR}, using a total of 43K real labels. Our network architecture and training hyperparameters are the same as \cite{Zhu2019VPLR}, which implies that the improvement is from self-training. This result is inspiring because it indicates the effectiveness of the self-training paradigm for semantic segmentation. We may not need a ultra large-scale labeled dataset to achieve good performance. Especially for autonomous driving, we have unlimited videos recorded during driving but not the resources to label them. With this self-training technique, we may generalize the model to various cities or situations with images alone. 

As shown in Table \ref{table:ratio}b, increasing the number of iterations by duplication is helpful, but not as good as using self-training ($79.1\%$ vs. $80.0\%$). In addition, the performance from duplication saturates at a certain number but self-training with pseudo labels continue to improve, which shows its potential to scale. Since a ratio of 7:1 between pseudo and real labels gives the best result, we will use this setting for the rest of our experiments unless otherwise stated.

\noindent \textbf{Generalizing to other students}
Self-training is model-agnostic. It is a way to increase the number of training samples, and improve the accuracy and robustness of model itself. Here we would like to show that the pseudo labels generated by our teacher model (DeepLabV3+ with ResNeXt50 backbone), can improve the performance of (1) a heavier model (DeepLabV3+ with WideResNet38 backbone \cite{Wu2016WideOrDeep}); (2) a fast model (FastSCNN \cite{Poudel2019FastSCNN}) and (3) another widely adopted segmentation model (PSPNet with ResNet101 backbone \cite{Zhao2017pspnet}). 

As shown in Table~\ref{table:otherstudents}, our self-training method can improve the student model irrespective of the backbones and network architectures, which demonstrates its great generalization capability. We want to emphasize again that for all three students, our results are not only better than their comparing baseline, but also outperforms the models pre-trained on Mapillary labeled data.
In addition, our trained FastSCNN model achieves an mIoU score of $72.5\%$ on the Cityscapes validation set, with only $1.1$M parameters.
% and inference can be run at a speed of $123.5$ fps. 
We believe this is a strong baseline for real-time semantic segmentation as compared to recent literature \cite{Nekrasov_2019_fastNASSeg,Li2019DFANet,Yu_BiSeNet_2018eccv}.

\begin{table}[t]
\parbox{.45\linewidth}{
\centering
\caption{Comparison of fast training schedules. coarse2fine+ using crop size warm-up is able to achieve 1.7x speed up without performance degradation}
\label{table:fast_main}
\scalebox{0.8}{
\begin{tabular}{l c c}
\hline\noalign{\smallskip}
  & w/o warm-up & w warm-up\\
\noalign{\smallskip}
\hline
\noalign{\smallskip}
Baseline & 80.0 & 80.0 \\
\noalign{\smallskip}
\hline
\noalign{\smallskip}
coarse2fine & 79.1 & 79.6 \\
fine2coarse & 79.4 & 79.8 \\
coarse2fine+ & 79.5 & \textbf{80.0} \\
fine2coarse+ & 79.4 & 79.9 \\
\hline
speed up & 1.8x & 1.7x \\
\hline
\end{tabular}
}
}
\hfill
\parbox{.5\linewidth}{
\centering
\caption{Results on the CamVid test set. Pre-train indicates the source dataset on which the model is trained}
\label{table:camvid}
\scalebox{0.7}{
\begin{tabular}{l c c}
    \hline\noalign{\smallskip}
     Method & Pretrain & mIoU ($\%$)\\
    \noalign{\smallskip}
    \hline
    \noalign{\smallskip}
	RTA \cite{Huang_2018_ECCV} & ImageNet & $62.5$    \\
	DFANet \cite{Li2019DFANet} 	& ImageNet  &  $64.7$    \\ 
	BiSeNet \cite{Yu_BiSeNet_2018eccv} 	 & ImageNet &   $68.7$    \\
	PSPNet \cite{Zhao2017pspnet}	  & ImageNet&   $69.1$    \\
	DenseDecoder \cite{Bilinski_2018_CVPR} & ImageNet	  &  $70.9$    \\
	SKDistill \cite{Liu2019KDSegmentation} & ImageNet	  &  $72.3$    \\
	VideoGCRF \cite{Chandra_2018_CVPR} 	& Cityscapes  &  $75.2$    \\ 
	SVCNet \cite{Ding2019Shape} 	& ImageNet  &  $75.4$    \\ 
	VPLR \cite{Zhu2019VPLR} 	& Cityscapes, Mapillary  &  $79.8$    \\ 
	\hline
	Ours 	 & Cityscapes &  $\textbf{81.6}$    \\
    \hline
\end{tabular}
}
}
\vspace{-4ex}
\end{table}

\noindent \textbf{Fast training schedules}
The surge in amount of training samples requires a faster training schedule. We introduce four of them: coarse2fine, fine2coarse, coarse2fine+ and fine2coarse+. As shown in Table \ref{table:fast_main}, our baseline uses a crop size of 800 throughout the training, and achieves $80.0\%$ mIoU. Next we simply switch to our proposed fast learning schedule, and obtain slightly worse performance with 1.8x speed up. However, our goal is to avoid the speed and accuracy trade-off. We find that a good initialization is important for semantic segmentation. Hence, we propose to warm-up the crop size in the early epochs. We use a large crop size of 800 in the first 20 epochs, and then switch to the fast training schedules. We can see that by using coarse2fine+ with crop size warm-up, our fast training schedule is able to match the performance of baseline with 1.7x speed up. Note that, the speed up is model dependent. We can enjoy 2x speed up when training with a heavier model using the WideResNet38 backbone.

\begin{table}[t]
	\begin{center}
		\caption{Per-class comparison with top-performing methods on the test set of Cityscapes. Our method outperforms all prior literature that only uses fine labels. In terms of fair comparison, our self-trained model achieves the same segmentation accuracy as a model pre-trained using Mapillary labeled data under identical settings \label{table:cs_sota}}
		\vspace{-2ex}
		\resizebox{1\columnwidth}{!}{%
			\begin{tabular}{  c | c  c  c  c  c c  c  c c  c  c c  c  c c  c  c c  c | c }
				\hline
				Method & road  & swalk & build. & wall & fence & pole & tlight & tsign & veg. & terrain & sky & person & rider & car & truck & bus & train & mcycle &  bicycle & mIoU   \\
				\hline
				DepthSeg \cite{Kong2018depthseg}  &  98.5  &  85.4  &  92.5  &  54.4  &  60.9  &  60.2  &  72.3   & 76.8   & 93.1   & 71.6   & 94.8  &  85.2  &  68.9   & 95.7  &  70.1  &  86.5  &  75.5   & 68.3   & 75.5   & 78.2 \\
				PSPNet  \cite{Zhao2017pspnet}       &  98.6 & 86.2 & 92.9 & 50.8 & 58.8 & 64.0 & 75.6 &  79.0 & 93.4 & 72.3 & 95.4 & 86.5 & 71.3 & 95.9 & 68.2 & 79.5 & 73.8 & 69.5 & 77.2 & 78.4        \\
				AAF  \cite{Ke2018AAF} &  98.5  & 85.6  & 93.0  & 53.8  & 58.9  & 65.9  & 75.0  & 78.4  & 93.7  & 72.4  & 95.6  & 86.4 &  70.5 &  95.9  & 73.9 &  82.7 &  76.9 &  68.7  & 76.4  & 79.1 \\
				PanoDeepLab \cite{Cheng2019DeepLabPano} &  98.7 & 87.2 &  93.6  & 57.7  & 60.8  & 70.8 &  78.0 &  81.2 &  93.8 &  74.1 &  95.7 &  88.2 &  76.4  & 96.0  & 55.3  & 75.1 &  79.6 &  72.1 &  74.0  &     $79.4$        \\
				DenseASPP \cite{Yang2018DenseASPP}  & 98.7   & 87.1  &  93.4  &  60.7  &  62.7   & 65.6   & 74.6  &  78.5  &  93.6  &  72.5  &  95.4  &  86.2   & 71.9   & 96.0   & 78.0  &  90.3   & 80.7   & 69.7   & 76.8  &  80.6 \\
				SPG  \cite{Cheng2019SPGNet}        & 98.8 & 87.6  & 93.8  & 56.5 &  61.9 &  71.9 &  80.0  & 82.1  & 94.1 &  73.5  & 96.1  & 88.7  & 74.9 &  96.5  & 67.3  & 84.8 &  81.8  & 71.1  & 79.4   &     $81.1$        \\
				% SeENet  &  98.7 &  87.3  & 93.7  & 57.1  & 61.8  & 70.5  & 77.6  & 80.9 &  94.0  & 73.5 &  95.9  & 87.5  & 71.6  & 96.3 &  76.4  & 88.0  & 79.9  & 73.0  & 78.5  & 81.2 \\
				BFP \cite{Ding2019BAFP}  & 98.7  & 87.0 &  93.5 &  59.8  & 63.4  & 68.9  & 76.8  & 80.9  & 93.7 &  72.8 &  95.5  & 87.0  & 72.1  & 96.0  & 77.6  & 89.0  & 86.9  & 69.2  & 77.6 &  81.4 \\
				DANet  \cite{Fu2018DANet}        &  98.6   &   86.1   &   93.5  &    56.1  &    63.3  &    69.7   &   77.3  &    81.3  &    93.9   &   72.9  &    95.7  &    87.3   &   72.9   &   96.2   &   76.8   &   89.4   &   86.5   &   72.2  &    78.2 &     $81.5$        \\
				HRNetv2  \cite{Sun2019HRNet}        &  $98.8$     &     $87.9$   &  $93.9$     &     $61.3$ &  $63.1$     &     $72.1$ &  $79.3$     &     $82.4$ &  $94.0$     &     $73.4$ &  $96.0$     &     $88.5$ &  $75.1$     &     $96.5$ &  $72.5$     &     $88.1$ &  $79.9$     &     $73.1$ &  $79.2$     &     $81.8$        \\
				ACFNet \cite{Zhang2019ACFNet}   &  98.7  & 87.1  & 93.9  & 60.2  & 63.9  & 71.1  & 78.6  & 81.5  & 94.0  & 72.9  & 95.9  & 88.1  & 74.1 &  96.5  & 76.6  & 89.3  & 81.5  & 72.1  & 79.2 &  81.8 \\
				EMANet  \cite{Li2019EMANet}        &  $98.7$     &     $87.3$   &  $93.8$     &     $63.4$ &  $62.3$     &     $70.0$ &  $77.9$     &     $80.7$ &  $93.9$     &     $73.6$ &  $95.7$     &     $87.8$ &  $74.5$     &     $96.2$ &  $75.5$     &     $90.2$ &  $84.5$     &     $71.5$ &  $78.7$     &     $81.9$        \\
				ACNet \cite{Fu2019ACNet} &  98.7 & 87.1 &  93.9  & 61.6  & 61.8  & 71.4 &  78.7 &  81.7 &  94.0 &  73.3 &  96.0 &  88.5 &  74.9  & 96.5  & 77.1  & 89.0 &  89.2 &  71.4 &  79.0  &     \secbest{82.3}        \\
				% AdapNet++  \cite{Zhao2017pspnet}        &  $98.7$     &     $86.9$   &  $93.5$     &     $58.4$ &  $63.7$     &     $67.7$ &  $76.1$     &     $80.5$ &  $93.6$     &     $72.2$ &  $95.3$     &     $86.8$ &  $71.9$     &     $96.2$ &  $77.7$     &     $91.5$ &  $83.6$     &     $70.8$ &  $77.5$     &     $81.3$        \\
				\hline
				Baseline   &  $98.7$     &     $86.7$   &  $93.6$     &     $59.0$ &  $63.1$     &     $68.6$ &  $77.1$     &     $80.4$ &  $94.1$     &     $73.7$ &  $96.0$     &     $87.5$ &  $73.0$     &     $96.2$ &  $73.2$     &     $85.6$ &  $86.5$     &     $70.4$ &  $77.1$     &     81.4        \\
				Mapillary-pretrained   &  $98.8$     &     $87.6$   &  $94.1$     &     $63.8$ &  $64.7$     &     $70.4$ &  $78.1$     &     $82.1$ &  $94.2$     &     $73.5$ &  $96.1$     &     $88.3$ &  $73.7$     &     $96.3$ &  $77.2$     &     $90.9$ &  $90.4$     &     $71.9$ &  $79.0$     &     \best{82.7}        \\
				Self-training   &  $98.8$     &     $87.8$   &  $94.0$     &     $61.7$ &  $64.9$     &     $71.6$ &  $78.6$     &     $82.2$ &  $94.2$     &     $74.2$ &  $96.1$     &     $88.4$ &  $74.3$     &     $96.5$ &  $76.7$     &     $90.1$ &  $90.0$     &     $72.3$ &  $79.1$     &     \best{82.7}        \\
				\hline
				% Ours   &  $98.8$     &     $87.8$   &  $94.0$     &     $61.7$ &  $64.9$     &     $71.6$ &  $78.6$     &     $82.2$ &  $94.2$     &     $74.2$ &  $96.1$     &     $88.4$ &  $74.3$     &     $96.5$ &  $76.7$     &     $90.1$ &  $90.0$     &     $72.3$ &  $79.1$     &     \best{82.7}        \\
				% \hline
				% % OCNetv4   &  $98.7$     &     $87.0$   &  $93.9$     &     $59.5$ &  $63.7$     &     $71.4$ &  $78.2$     &     $82.2$ &  $94.0$     &     $73.0$ &  $95.8$     &     $88.0$ &  $73.0$     &     $96.4$ &  $78.0$     &     $90.9$ &  $83.9$     &     $73.8$ &  $78.9$     &     $84.5$ \\
				% panoptic   &  $98.7$     &     $87.0$   &  $93.9$     &     $59.5$ &  $63.7$     &     $71.4$ &  $78.2$     &     $82.2$ &  $94.0$     &     $73.0$ &  $95.8$     &     $88.0$ &  $73.0$     &     $96.4$ &  $78.0$     &     $90.9$ &  $83.9$     &     $73.8$ &  $78.9$     &     $84.2$ \\
				% vplr   &  $98.7$     &     $87.0$   &  $93.9$     &     $59.5$ &  $63.7$     &     $71.4$ &  $78.2$     &     $82.2$ &  $94.0$     &     $73.0$ &  $95.8$     &     $88.0$ &  $73.0$     &     $96.4$ &  $78.0$     &     $90.9$ &  $83.9$     &     $73.8$ &  $78.9$     &     $83.5$ \\
				% % gald   &  $98.7$     &     $87.0$   &  $93.9$     &     $59.5$ &  $63.7$     &     $71.4$ &  $78.2$     &     $82.2$ &  $94.0$     &     $73.0$ &  $95.8$     &     $88.0$ &  $73.0$     &     $96.4$ &  $78.0$     &     $90.9$ &  $83.9$     &     $73.8$ &  $78.9$     &     $83.3$ \\
				\hline
			\end{tabular}
		}
	\end{center}
	\vspace{-5ex}
\end{table}

\noindent \textbf{Comparison to state-of-the-art}
We compare our self-training method to recent literature on the test set of Cityscapes. For the test submission, we train our model using the best recipe suggested above, with several modifications. We use WideResNet38 \cite{Wu2016WideOrDeep} as the backbone and adopt a standard multi-scale strategy following \cite{Zhao2017pspnet,Chen2018deeplabv3plus} to perform inference on multi-scaled (0.5, 1.0 and 2.0), left-right flipped and overlapping-tiled images. 
% More details can be found in the supplementary materials.
As we can see in Table \ref{table:cs_sota}, our self-training method achieves an mIoU of $82.7\%$ using only the Cityscapes fine annotations, outperforming all prior methods that use the same training data. Note that most prior approaches also adopt other effective techniques for semantic segmentation, such as attention mechanism \cite{Fu2018DANet,Fu2019ACNet,Zhang2019ACFNet}, improved boundary handling \cite{Ding2019BAFP,Zhu2019VPLR}, better network architecture \cite{Sun2019HRNet,Li2020dynamicRoute} or multi-tasking framework \cite{Kong2018depthseg,Cheng2019DeepLabPano}. Our self-training framework is orthogonal to all these techniques, and can incorporate them to further improve the performance. In addition, we even outperform some recent approaches using external labeled data, such as InPlaceABN \cite{Bulo2018inplaceABN}, Auto-DeepLab-L \cite{Liu2019autodeeplab}, SSMA \cite{Valada2019SSMA} and DPC \cite{Chen2018SearchSegDPC}, etc.

% Some of those methods use attention mechanism \cite{Fu2018DANet,Fu2019ACNet,Zhang2019ACFNet}, improved boundary handling \cite{Ding2019BAFP}, better network architecture \cite{Sun2019HRNet} or multi-tasking framework \cite{Kong2018depthseg,Cheng2019DeepLabPano}. our method 

\begin{figure}[t]
\centering
\includegraphics[width=1.0\linewidth]{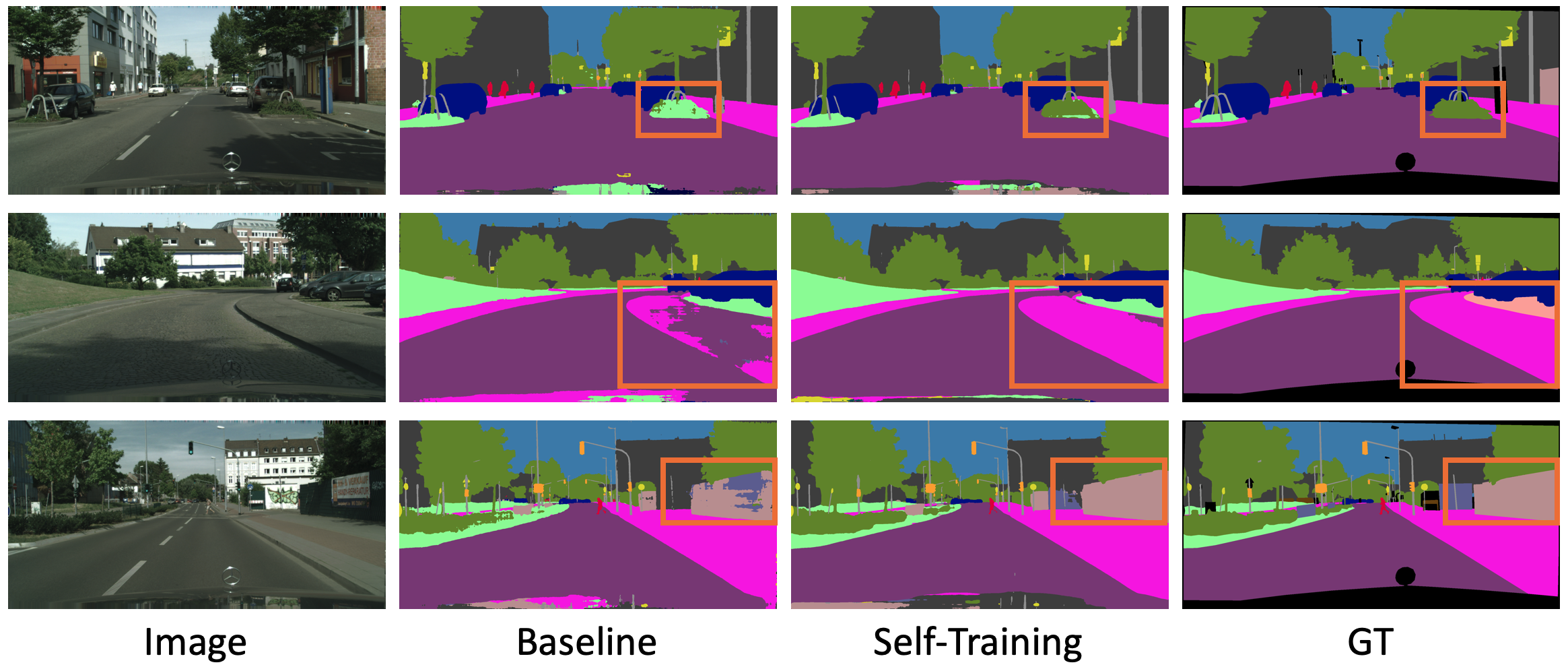}
\vspace{-4ex}
\caption{Visual comparisons on Cityscapes. We demonstrate that self-training can effectively handle class confusion, such as between tree and vegetation (row 1), road and sidewalk (row 2), wall and fence (row 3).}
\label{fig:visualization}
\vspace{-2ex}
\end{figure}

In order to show the contribution from self-training alone, we perform a fair comparison in the bottom of Table \ref{table:cs_sota}. We design three training settings with the same network architecture, (a) Baseline: we only use the fine annotations; (b) Mapillary pre-trained: we use Mapillary labeled data to pre-train the model and then finetune it on Cityscapes; and (c) Self-training: our proposed method using only the fine annotations from Cityscapes and pseudo labels generated from Mapillary. As we can see, our self-training approach has the same segmentation accuracy as the model pre-trained using Mapillary labeled data ($82.7\%$). To clarify, we cannot do joint training when using Mapillary labeled data because Mapillary has different classes. We also show several visual examples in Fig.~\ref{fig:visualization}, and demonstrate that self-training can handle class confusion better than the baseline model. In conclusion, we demonstrate the effectiveness of our proposed self-training framework for semantic segmentation, achieving state-of-the-art performance while requiring significantly less supervision. 

Lastly, we also experimented with several modifications, for instance comparison between using hard (a one-hot distribution) or soft (a continuous distribution) labels, single-loop or multi-loop teacher-student, joint training or pre-training followed by finetuning. In terms of labels, our experiments show that using hard labels in general performs better than using soft labels. This intuitively makes sense because dense prediction problems favor hard labels \cite{He2019BagTricks}. In terms of teacher-student iteration, we do not observe improvement using more loops of self-training for semantic segmentation, i.e., putting back the student as teacher and train another student model. In terms of training, we try a conventional setting \cite{Yalniz2019BillionSemi} to first pre-train on pseudo labels and then finetune it on real labels, but we observe worse performance than joint training with all the labels. Detailed comparisons on these ablation studies can be found in the Appendix. 

\subsection{Finetuning on Other Datasets}
\label{subsec:kitti_camvid}
In this section, we would like to show that our student model serves as a good initialization for finetuning on other driving scene segmentation datasets, such as CamVid and KITTI. To be specific, we take our well-trained student model (DeepLabV3+ with WideResNet38 backbone), and finetune it on target datasets to report the performance.

For CamVid, we only report single-scale evaluation scores on its test set for fair comparison to previous approaches. As can be seen in Table \ref{table:camvid}, we achieve state-of-the-art performance, an mIoU of $81.6\%$. We want to point out that the previous best method \cite{Zhu2019VPLR} uses a model pre-trained on both Cityscapes and Mapillary labeled data as initialization to finetune on CamVid. We outperform it by $1.8\%$. This result is encouraging because generalizing to other datasets is usually more meaningful \cite{He2019MomentumContrast}. In addition, it supports our claim that a model learned using the self-training framework is more robust.

We also achieve promising results on the KITTI leaderboard, with an mIoU of $71.41\%$. Given limited space, detailed results can be found in the Appendix.

\begin{figure}[t]
\vspace{-0.5cm}
    \begin{center}
        \input{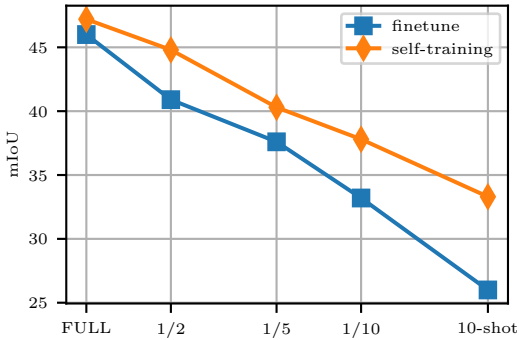}
    \end{center}
    \vspace{-0.5cm}
    \caption{Experiments on cross-domain generalization from Cityscapes to Mapillary. Full: full training set of Mapillary (18K  samples). 1/2, 1/5 and 1/10 means we use 1/2, 1/5 and 1/10 of the full training set. 10-shot: 10 samples per class}
    \label{fig:few_shot}
    \vspace{-0.5cm}
\end{figure}

\subsection{Cross-Domain Generalization with New Categories}
\label{subsec:mapillary}

Here, we first describe the task of cross-domain generalization, and then show the robustness of our self-training method. We set Cityscapes dataset as the source domain, and Mapillary dataset as the target domain. We choose Mapillary due to its large size and variability. In addition, Mapillary has many more semantic categories than Cityscapes, 66 compared to 19. Hence, most classes are considered new, never seen by the model trained on the source domain, which makes the problem very challenging. We randomly select 1/10, 1/5 and 1/2 of the original Mapillary training set to form different evaluation scenarios.

We compare two methods, (1) conventional finetuning approach: a model trained on Cityscapes fine annotations, and then finetuned on the given Mapillary labeled samples; (2) our self-training approach: a model trained on both fine annotations from Cityscapes and pseudo labels from Mapillary, and then finetuned on the given Mapillary labeled samples. For simplicity, we do not use OHEM loss and other non-default training options, to better demonstrate the contributions from model initialization. Our goal is to see whether self-training can help improve a model's generalization capability across domains.

We can see the results in Fig. \ref{fig:few_shot}. First, the model trained using our self-training framework achieves the best performance across all scenarios compared to the finetuning approach. This indicates that the self-training method can largely reduce the human labeling effort when generalizing to new domains. Second, in terms of using the full training set, our approach also performs better. This maybe due to the fact that self-training serves as effective data augmentation and provides a better model initialization. Third, given fewer and fewer training samples, our method starts to reveal its robustness. For example, using only 1/10 of the training set, we are able to achieve $37.8\%$ mIoU, which even outperforms the finetuning model using 1/5 of the data. We can see that the gap between the finetuning model and our self-training model becomes larger.

We would like to push this task even further to the few-shot case, where we only have $10$ training samples per class. As can be seen in Fig. \ref{fig:few_shot}, our method is still able to achieve decent performance of $33.3\%$ mIoU. We significantly outperform the finetuning approach, an improvement of $7.3\%$ ($26.0\% \shortrightarrow 33.3\%$). In addition, our performance using 10 samples per class is even better than the finetuning approach using 1/10 of the full training set (30 samples per class). Hence, the results suggest that our self-training technique can generalize well to various locations with just images and a few annotations from the new scene. Another cross-domain generalization experiment, from Cityscapes to BDD100K, can be found in the Appendix. We will release the data splits used in our cross-domain generalization experiments for fair comparison.

\vspace{-2ex}
\section{Conclusion}
\label{sec:conclusion}
In this work, we introduce a self-training framework for driving scene semantic segmentation. The self-training method can leverage a large number of unlabeled data to improve both accuracy and robustness of the segmentation model. 
Together with our proposed training techniques, we achieve state-of-the-art performance on three driving scene benchmark datasets, Cityscapes, CamVid and KITTI, while requiring significantly less supervision. We also demonstrate that our self-training method works well on a challenging cross-domain generalization task. Even with 10 labeled samples per class, we show that our model is able to achieve decent segmentation accuracy generalizing from Cityscapes to Mapillary. Lastly, we propose a fast training schedule, which is a general technique to speed up model learning by 2x without performance degradation.

\bibliographystyle{splncs04}
\bibliography{egbib}

% \clearpage

\appendix
\section*{Appendix}

We will provide more details about the results in the main paper, and show more visualizations. To be specific, we first discuss three alternative design choices of our self-training framework in Appendix.~\ref{sec:ablation}. Then we show visualizations of the pseudo labels, successful predictions and failure cases in Appendix.~\ref{sec:visualizations}. Furthermore, we present two experiments on using a better teacher and cross-domain generalization in Appendix.~\ref{sec:generalization}. In the end, we provide more details of our KITTI results in Appendix.~\ref{sec:kitti}.

\section{Ablation Studies}
\label{sec:ablation}
We have indicated at the end of Sec. 4.3 of the main paper that we could only briefly describe the ablation studies due to space limitations. Here, we discuss more details of the ablation studies. The baseline is a DeepLabV3+ model with ResNeXt50 backbone, the same as in the main paper. The ratio between pseudo labels to real labels defaults to 7:1. We still run the same experiment five times and report the average mIoU.

\subsection{Hard vs soft labels}
\label{subsec:hard}
The generated labels could be either hard or soft. Hard means it is a one-hot distribution. We can think of them as ground truth labels and use standard cross-entropy loss to start training. Soft means it is a continuous distribution (i.e., each label has an associated probability). We need to save the probabilities from the teacher's predictions as supervision and use sparse cross-entropy loss to train the student model. 

Table~\ref{table:soft_hard} shows that the model does better with hard labels than soft labels. This observation agrees with \cite{He2019BagTricks} that dense prediction problems favor hard labels. A potential explanation is that soft labels may cause ambiguity around object boundaries, which is harmful for semantic segmentation. This also indicates the difference between image classification and semantic segmentation, because soft labels are usually preferred in the image classification domain \cite{Yalniz2019BillionSemi,Xie2019NoisyStudent}.

\subsection{Joint learning vs pre-training}
\label{subsec:joint}
Once we have a large set of pseudo labels, we train our student model using a joint learning process with both human-annotated real labels and teacher-generated pseudo labels. Here, we would like to compare it with a conventional alternative following \cite{Yalniz2019BillionSemi}: pre-training on the large set of pseudo labels first and then finetuning on the real labels. 

Table~\ref{table:joint} shows that joint learning outperforms the conventional pipeline using pre-training and then finetuning. A potential explanation is that joint training with both real and pseudo labels serves as an effective data augmentation to regularize the model learning. 

% Here, we investigate their impact on the final performance.
\subsection{Single-loop or multi-loop}
\label{subsec:loop}
Teacher-student learning could be iterative which means we can use the student as teacher, generate more accurate pseudo labels and then retrain another student model. Here, we use more loops of self-training to see if helps semantic segmentation. 

As seen in Table~\ref{table:loop}, using a single-loop of teacher-student training is able to achieve promising results ($80.0\%$). 2-loop obtains slightly worse results ($79.9\%$), and 3-loop is  slightly better ($80.2\%$). In terms of a good trade-off between accuracy and resources, we only perform a single iteration of teacher-student for all experiments.

\begin{table}[t]
\parbox{.45\linewidth}{
\centering
\caption{Hard vs soft labels}
\label{table:soft_hard}
\scalebox{0.8}{
\begin{tabular}{l  c}
\hline\noalign{\smallskip}
  & mIoU ($\%$) \\
\noalign{\smallskip}
\hline
\noalign{\smallskip}
Baseline & 78.1 \\
\noalign{\smallskip}
\hline
\noalign{\smallskip}
Soft Pseudo & 79.6 \\
Hard Pseudo (ours) & \textbf{80.0} \\
\hline
\end{tabular}
}
}
\hfill
\parbox{.45\linewidth}{
\centering
\caption{Joint learning vs pre-training}
\label{table:joint}
\scalebox{0.8}{
\begin{tabular}{l c c}
    \hline\noalign{\smallskip}
  & mIoU ($\%$) \\
\noalign{\smallskip}
\hline
\noalign{\smallskip}
Baseline & 78.1 \\
\noalign{\smallskip}
\hline
\noalign{\smallskip}
Pre-training then finetuning & 79.2 \\
Joint learning (ours) & \textbf{80.0} \\
    \hline
\end{tabular}
}
}
% \vspace{-4ex}
\end{table}

\begin{table}[t]
\parbox{.45\linewidth}{
\centering
\caption{Single-loop vs multi-loop }
\label{table:loop}
\scalebox{0.8}{
\begin{tabular}{l  c}
\hline\noalign{\smallskip}
  & mIoU ($\%$) \\
\noalign{\smallskip}
\hline
\noalign{\smallskip}
Baseline & 78.1 \\
\noalign{\smallskip}
\hline
\noalign{\smallskip}
1-loop & 80.0 \\
2-loop & 79.9 \\
3-loop & \textbf{80.2} \\
\hline
\end{tabular}
}
}
\hfill
\parbox{.45\linewidth}{
\centering
\caption{Fast training schedule}
\label{table:fast_appendix}
\scalebox{0.8}{
\begin{tabular}{l l c}
    \hline\noalign{\smallskip}
 Network & Backbone  & Speed up \\
\noalign{\smallskip}
\hline
\noalign{\smallskip}
FastSCNN  \quad  \quad  & - & 1.5x \\
DeepLabV3+  \quad  \quad & ResNeXt50 & 1.7x \\
PSPNet  \quad  \quad & ResNet101 & 1.9x \\
DeepLabV3+  \quad  \quad & WideResNet38 & \textbf{2x} \\
\hline
\end{tabular}
}
}
% \vspace{-4ex}
\end{table}

\subsection{Model-dependent speed up}
\label{subsec:speed}
Recall from Sec. 3.3 in the main paper, that the speed up using our proposed fast training schedule is model-dependent. Here, we show the detailed speed up information for various models.

As seen in Table~\ref{table:fast_appendix}, larger models tend to benefit more from the fast training schedule. For example, we  achieve 2x speed up when training on a DeepLabV3+ model with WideResNet38 backbone. This is because when the model is bigger, the time spent on network computation  dominates the training time. If we reduce the crop size, we  save a lot of computation. 

% \subsection{details about centroid sampling}

\begin{figure}[t]
\centering
\includegraphics[width=1.0\linewidth]{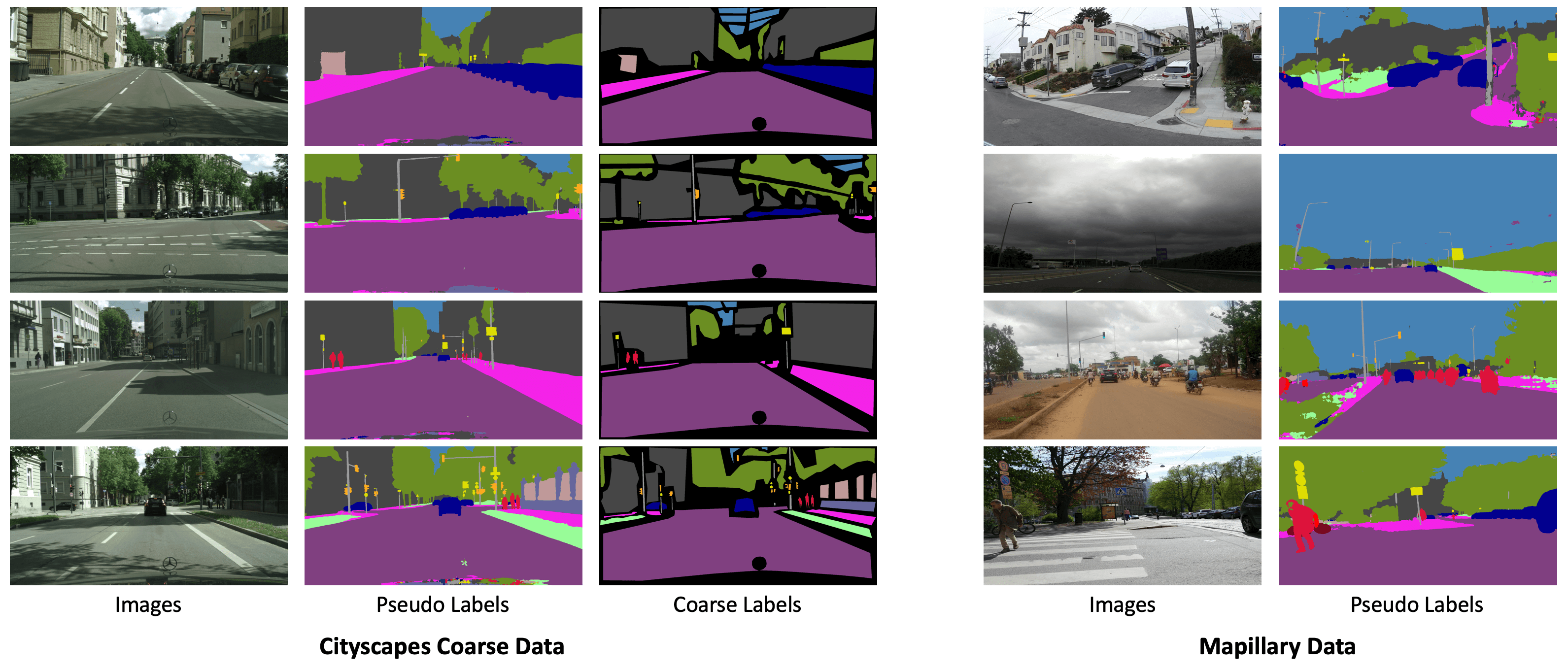}
% \vspace{-4ex}
\caption{Visualizations of our teacher-generated pseudo labels. Left: on Cityscapes coarse images. We can see our pseudo labels have higher quality than human-labeled coarse annotations based on polygon. Black regions in coarse annotations represent background class. Right: on Mapillary images. Our teacher model is able to provide reasonable segmentation predictions despite the large domain gap.}
\label{fig:pseudo_labels}
\vspace{-2ex}
\end{figure}

\section{Visualizations}
\label{sec:visualizations}
Here, we first show several visualizations of our teacher-generated pseudo labels in Appendix.~\ref{subsec:pseudo}. Then we provide some successful predictions and failure cases on Cityscapes dataset in Appendix.~\ref{subsec:cityscapes_appendix}.

% \vspace{-2ex}
\subsection{Visualizations of pseudo labels}
\label{subsec:pseudo}
As mentioned in Sec. 3.1 of the main paper, we  show more visualizations of pseudo labels here in Fig.~\ref{fig:pseudo_labels}. 

First, we look at our generated pseudo labels on Cityscapes coarse images (left column Fig.~\ref{fig:pseudo_labels}). Compared to the coarse labels annotated by humans (i.e., polygons), our pseudo labels have higher quality, such as sharper boundaries, correct predictions, etc. For example on row 3, our pseudo labels successfully capture several persons on the right of the sidewalk, while the coarse annotations completely ignore them.

Then, we show our generated pseudo labels on Mapillary images. The Mapillary dataset is collected worldwide, and includes different seasons, time of the day, traffic etc. Our teacher model is able to provide reasonable predictions on these challenging situations, such as uphill road (row 1), cloudy weather (row 2) and other countries (row 3 and 4). 

Despite some erroneous predictions, the quality of our teacher-generated pseudo labels are in general good. This is part of the reason our self-training method works, because the student model won't learn well if the pseudo labels contain too much noise. 

\begin{figure}[t]
\centering
\includegraphics[width=1.0\linewidth]{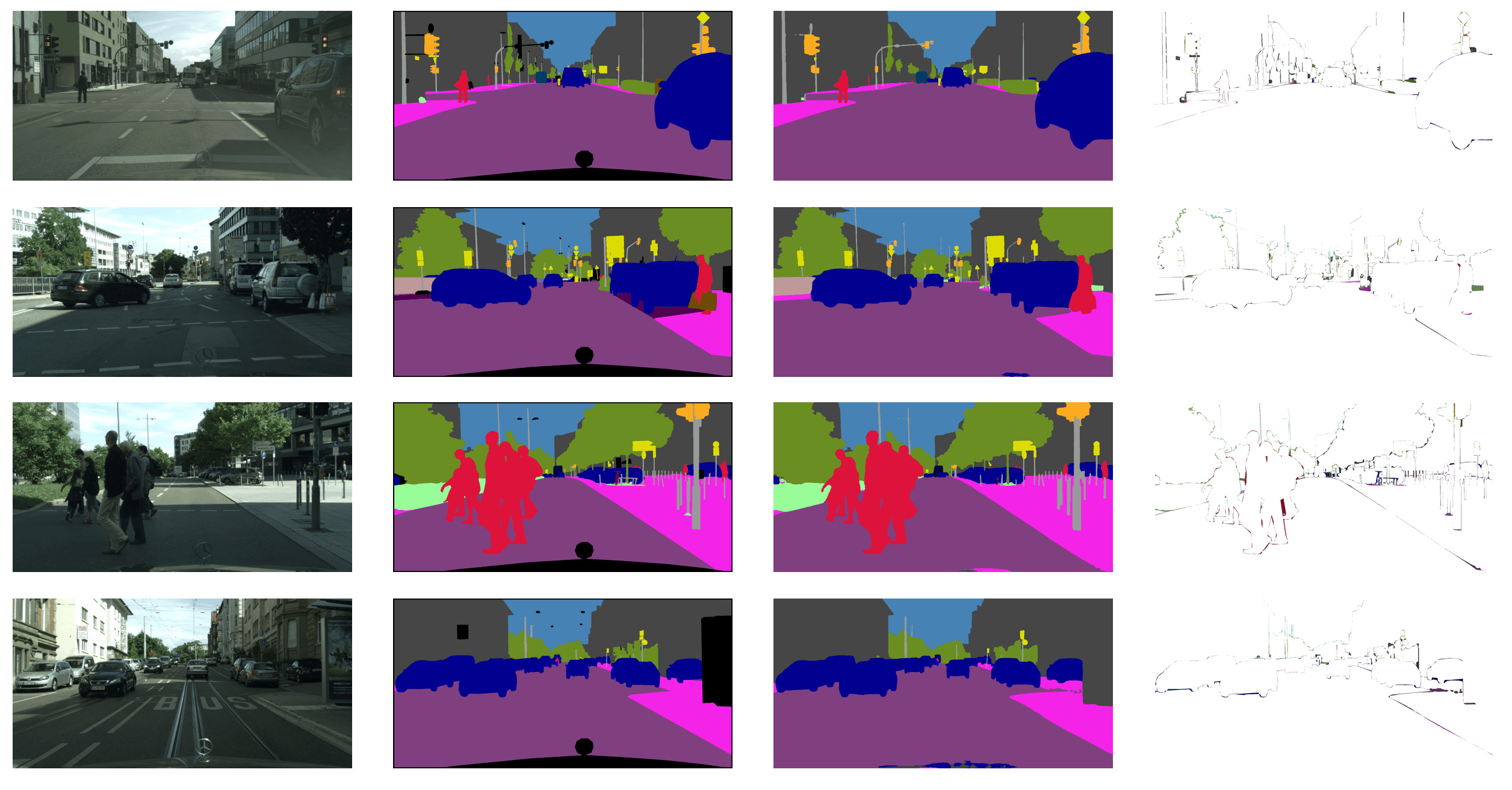}
% \vspace{-4ex}
\caption{Successful predictions on Cityscapes dataset. From left to right: image, ground truth, our prediction and their differences. We can see that our model is able to assign correct semantic labels to each pixel except the object boundaries due to challenges such as annotation ambiguity. Black regions in ground truth represent background class}
\label{fig:cs_good}
\vspace{-2ex}
\end{figure}

\begin{figure}[!ht]
\centering
\includegraphics[width=1.0\linewidth]{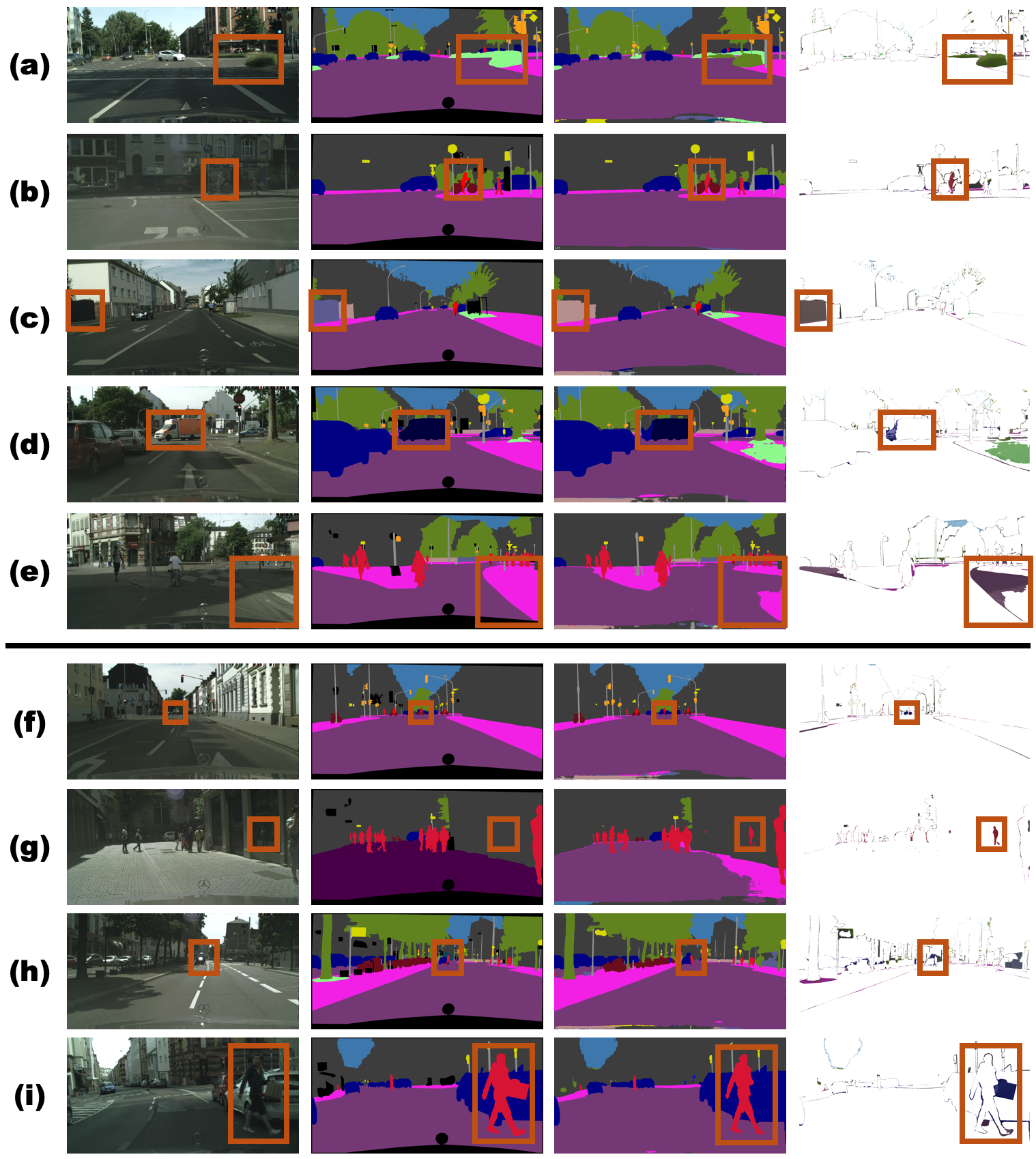}
% \vspace{-4ex}
\caption{Visualizations of failure cases. Top: five common scenarios of class confusion. From rows (a) to (e), our model has difficulty in segmenting: (a) terrain and vegetation, (b) person and rider, (c) wall and fence, (d) car and truck, (e) road and sidewalk. Bottom: challenging situations, (f) objects overlapping, (g) reflection in the mirror, (h) objects far away with strong illumination, and (i) annotation ambiguity. }
\label{fig:cs_fail}
\vspace{-4ex}
\end{figure}

% \vspace{-2ex}
\subsection{Visualization on Cityscapes}
\label{subsec:cityscapes_appendix}
We first show several visualizations of our successful predictions in Fig.~\ref{fig:cs_good}. We see that our model is able to handle small objects, crowded scenes, complicated lighting, etc. Our predictions are accurate and sharp, as demonstrated by the difference images in the rightmost column. We can assign correct semantic labels to each pixel except the object boundaries due to challenges such as annotation ambiguity.

Then we show visualizations of the failure cases of our model on Cityscapes dataset in Fig.~\ref{fig:cs_fail}. We show five common scenarios of class confusion. From rows (a) to (e), our model has difficulty in segmenting: (a) terrain and vegetation, (b) person and rider, (c) wall and fence, (d) car and truck, (e) road and sidewalk. Some of these situations are very challenging. For example, the wall in row (c) has holes, which by definition should be a fence. 

In addition, we show four even challenging scenarios from rows (f) to (i). In Fig.~\ref{fig:cs_fail} (f), the bicycle is overlapping with a car. It is hard to correctly tell them apart from such a long distance. In Fig.~\ref{fig:cs_fail} (g), our model predicts a reflection in the mirror as a person. This is an interesting result because our prediction should be considered correct in terms of appearance without reasoning about context. In Fig.~\ref{fig:cs_fail} (h), it is very hard to tell whether it is a car or bus when the object is far away, especially when there is strong illumination. In Fig.~\ref{fig:cs_fail} (i), this is an ambiguous situation because the handbag is neither a person or car. It should be a background class.

\setlength{\tabcolsep}{5pt}
\begin{table}[t]
\begin{center}
\caption{A better teacher model helps the learning of student model. $\Delta$ indicates the improvement from changing to a better teacher}
\vspace{2ex}
\label{table:better_teacher}
\resizebox{0.6\columnwidth}{!}{%
\begin{tabular}{l l l c c }
\hline\noalign{\smallskip}
 & Network & Backbone  & mIoU ($\%$) & $\Delta$ ($\%$)\\
\noalign{\smallskip}
\hline
\noalign{\smallskip}
Teacher A & DeepLabv3+  & ResNeXt50  & 78.1 & \\
\noalign{\smallskip}
\hline
\noalign{\smallskip}
  & DeepLabV3+  & ResNeXt50  & 80.0 & \\
  & DeepLabV3+  & WideResNet38  & 82.2 & \\
  & FastSCNN  &  -  & 72.5 & \\
  & PSPNet  &  ResNet101 & 79.9 & \\
\noalign{\smallskip}
\hline
\noalign{\smallskip}
Teacher B & DeepLabv3+  & WideResNet38  & 80.5 & +2.4\\
\noalign{\smallskip}
\hline
\noalign{\smallskip}
  & DeepLabV3+  & ResNeXt50  & 80.4 & +0.4\\
  & DeepLabV3+  & WideResNet38  & 82.7 & +0.5\\
  & FastSCNN  &  -  & 72.8 & +0.3\\
  & PSPNet  &  ResNet101 & 80.1 & +0.2\\
\hline
\end{tabular}
}
\end{center}
% \vspace{-3ex}
\end{table}

\begin{table}[t]
\begin{center}
\caption{Generalizing from Cityscapes to BDD100K. 10-shot means we only use 10 samples per class from BDD100K to train the model (i.e., a total of 200 training samples). Full means we use the full BDD100K dataset (i.e., a total of 7K samples)}
\label{table:cross}
\resizebox{0.32\columnwidth}{!}{%
\begin{tabular}{l c c}
    \hline\noalign{\smallskip}
  & 10-shot  & full \\
\noalign{\smallskip}
\hline
\noalign{\smallskip}
Finetuning  \quad  \quad  & 51.4 & 62.3 \\
Self-training  \quad  \quad & 58.9 & 65.7 \\
\hline
\end{tabular}
}
\end{center}
\vspace{-2ex}
\end{table}

\vspace{-2ex}
\section{Generalization}
\label{sec:generalization}

\subsection{Does a better teacher help?}
\label{subsec:better_teacher}
In the main paper, we always use a DeepLabV3+ model with ResNeXt50 backbone as the teacher model, given its good accuracy and speed trade-off. However, a straightforward question arises, does a better teacher help? 

Here, we use a better model (DeepLabV3+ network with WideResNet38 backbone) as our teacher, we call it teacher B. Its performance on the validation set of Cityscapes is $80.5\%$, higher than $78.1\%$ of the old teacher which we call teacher A. We will use teacher B to generate pseudo labels and compare to Table 3 in the main paper to answer the question.

As can be seen in Table~\ref{table:better_teacher}, a better teacher indeed helps. The performance of all the student models improve. However, the improvements range from 0.2 to 0.5, not significant compared to the performance gap between teacher A and teacher B ($78.1\%$ vs $80.5\%$).

\subsection{Cross-domain generalization: from Cityscapes to BDD100K}
\label{subsec:cross}
% \vspace{-1ex}
As mentioned in Sec. 3.4 of the main paper, generalizing a trained model to other domains (i.e, datasets or locations) given limited supervision is one of the motivations and contributions of our work. 

We have done an experiment, generalizing from a model trained on Cityscapes to Mapillary, in Sec. 4.5 of the main paper. We demonstrate that our model can generalize to other domains with new categories, given a few labeled data.
% It is a challenging situation since Mapillary has 66 classes, much more than the 19 classes in Cityscapes. 
Here, we perform another experiment, generalizing from a model trained on Cityscapes to BDD100K \cite{BDD100K}. Despite the fact that BDD100K has the same number of classes (19) as Cityscapes, this is a challenging situation because the data of BDD100K is collected in United States, which has a big domain gap compared to the data from Cityscapes. 

As seen in Table~\ref{table:cross}, our self-training method outperforms the conventional finetuning approach by a large margin. We want to emphasize that even using 10 samples per class from BDD100K (i.e., a total of 200 training samples), our model can achieve an mIoU of $58.9\%$, close to the finetuning approach using the full dataset ($62.3\%$ by using 7K training samples). This result strongly indicates the effectiveness of our method's generalization capability.

\begin{figure}[t]
\centering
\includegraphics[width=1.0\linewidth]{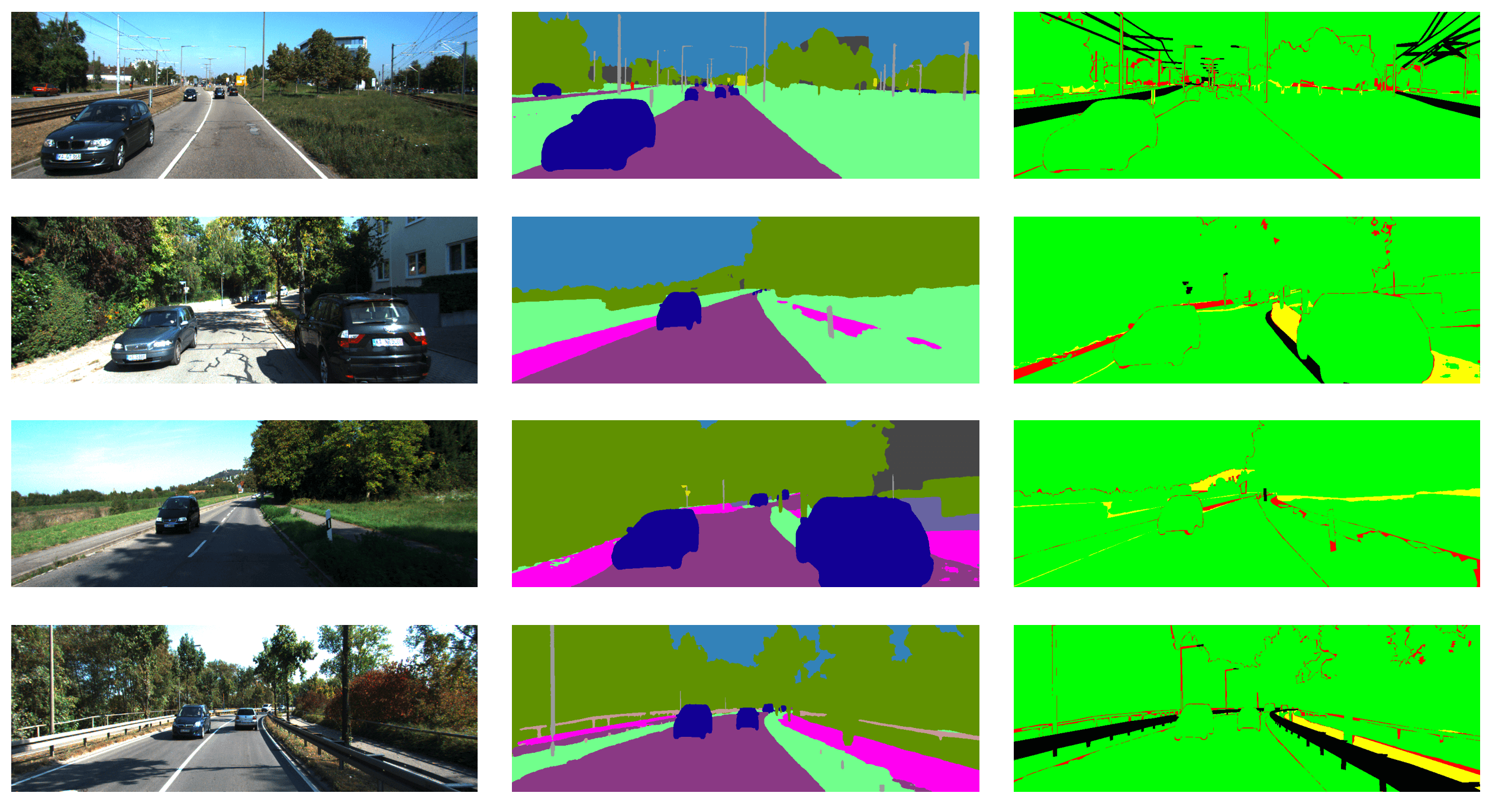}
\vspace{-2ex}
\caption{Successful predictions on KITTI dataset. From left to right: image, our prediction and the difference between our prediction and ground truth. Note that we do not have the ground truth for the test set. The difference images are provided by the KITTI official evaluation server. Black regions represent background class}
\label{fig:kitti_good}
\vspace{-2ex}
\end{figure}

\setlength{\tabcolsep}{4pt}
\begin{table}[t]
\begin{center}
	\caption{Results on KITTI test set. Pre-train indicates the source dataset on which the model is trained. I: ImageNet, C: Cityscapes, M: Mapillary and V: Cityscapes video \label{table:kitti}}
		\vspace{-3ex}
		\resizebox{0.7\columnwidth}{!}{%
\begin{tabular}{l c c c c c }
                \hline\noalign{\smallskip}
                Method & Pre-train & IoU class & iIoU class & IoU category & iIoU category \\
                \noalign{\smallskip}
                \hline
                \noalign{\smallskip}
                APMoE$\_$seg    \cite{kong2019pag} & I & $47.96$	& $17.86$	& $78.11$	& $49.17$ \\
				SegStereo \cite{yang2018segstereo}   & C & $59.10$	& $28.00$	& $81.31$	& $60.26$ \\
				AHiSS \cite{meletis2018AHiSS}	 & C, M & $61.24$	& $26.94$	& $81.54$	& $53.42$ \\
				LDN2 \cite{Ivan_ladderLDN_iccvw2017}	 & C, M & $63.51$	& $28.31$	& $85.34$	& $59.07$ \\
				MapillaryAI \cite{Bulo2018inplaceABN}	 & C, M & $69.56$	& $43.17$	& $86.52$	& $68.89$  \\
				VPLR \cite{Zhu2019VPLR}	 & C, V, M & \best{72.83}	&  \best{48.68}	& \best{88.99}	& \best{75.26} \\
				\hline
				Ours 	 & C & \secbest{71.41} & \secbest{46.09} & \secbest{88.28}  & \secbest{72.64}    \\
                \hline
            \end{tabular}
    }
\end{center}
\vspace{-2ex}
\end{table}

\vspace{-1ex}
\section{KITTI Results}
\label{sec:kitti}
Recall from Sec. 4.4 in the main paper, that we also achieved promising results on the KITTI leaderboard, with an mIoU of $71.41\%$ ranking 2nd. Here, we report the detailed results on the test set in Table~\ref{table:kitti}.

We would like to emphasize that our model is only pre-trained on Cityscapes fine annotations (about 3K samples), while other approaches \cite{meletis2018AHiSS,Ivan_ladderLDN_iccvw2017,Bulo2018inplaceABN,Zhu2019VPLR} use external training data, such as Mapillary Vista dataset and Cityscapes coarse annotations (for a total of about 43K samples). VPLR \cite{Zhu2019VPLR}, the top performer, also uses Cityscapes video data to help regularize the model training. For fair comparison in terms of training data usage, our method uses the same training data with SegStereo \cite{yang2018segstereo} but outperforms it by $12.3\%$. 

Qualitatively, we show several visualizations of our successful predictions in Fig.~\ref{fig:kitti_good}. These images are of the test set and the difference images are provided by KITTI official evaluation server.

\end{document}